\documentclass[letterpaper, 10pt, conference]{ieeeconf}  
\IEEEoverridecommandlockouts                              % This command is only needed if 
\usepackage{units}
\usepackage{bm}
\usepackage{float}
\usepackage[font=small,skip=7pt]{caption}
\usepackage[ruled,vlined,noend]{algorithm2e}
\usepackage{hyperref}
\usepackage{microtype}

\usepackage{epsfig}
\newlength\fwidth

\usepackage{mathtools}

\usepackage{xcolor}
\usepackage{siunitx}
\usepackage{multirow}
\usepackage{epstopdf}
\usepackage{amsmath} % assumes amsmath package installed
\usepackage{amssymb}  % assumes amsmath package installed
\usepackage{mathtools}
\DeclareGraphicsExtensions{.pdf,.eps}
\DeclareMathOperator*{\argmin}{arg\,min}
\usepackage[acronym,nomain,nonumberlist]{glossaries}
\newcommand{\N}{\mathbb{N}}
\newcommand{\R}{\mathbb{R}}
\newcommand\smashunderbracket[3][0.5pt]{%
	\vphantom{#2}%
	\smash{\underbracket[#1]{#2}_{#3}}%
}

\newacronym{mav}{MAV}{Micro Aerial Vehicle}
\newacronym{nmpc}{NMPC}{Nonlinear Model Predictive Control}
\newacronym{mpc}{MPC}{Model Predictive Control}
\newacronym{panoc}{PANOC}{Proximal Averaged Newton-type method for Optimal Control}

\usepackage{url}
\newcommand*{\smallmat}[1]
  {\left[\begin{smallmatrix}#1\end{smallmatrix}\right]}

\title{\LARGE \bf A Scalable Distributed Collision Avoidance Scheme for \\Multi-agent UAV systems %\thanks{This work has been partially funded by the European Unions Horizon 2020 Research and Innovation Programme under the Grant Agreement No. 730302 SIMS.}
}

  \author{Bj\"orn Lindqvist$^1$, Pantelis Sopasakis$^2$ and George Nikolakopoulos$^1$.
  
  \thanks{$^{1}$The authors are with the Robotics and AI Team, Department of Computer, Electrical and Space Engineering, Lule\r{a} University of Technology, Lule\r{a} SE-97187, Sweden}%
  \thanks{$^{2}$The author is with the School of Electronics, Electrical Engineering and Computer Science (EEECS), Queen's University Belfast and Centre for Intelligent Autonomous Manufacturing Systems (i-AMS), United Kingdom}%
  \thanks{Corresponding Author's email: \texttt{bjolin@ltu.se} }}
\begin{document}
\captionsetup{font=footnotesize}
\maketitle
\thispagestyle{empty}
\pagestyle{empty}
%%%%%%%%%%%%%%%%%%%%%%%%%%%%%%%%%%%%%%%%%%%%%%%%%%%%%%%%%%%%%%%%%%%%%%%%%%%%%%%%
\begin{abstract}
In this article we propose a distributed collision avoidance scheme for multi-agent unmanned aerial vehicles (UAVs) based on nonlinear model predictive control (NMPC), where other agents in the system are considered as dynamic obstacles with respect to the ego agent. 
Our control scheme operates at a low level and commands roll, pitch and thrust signals at a high frequency,
each agent broadcasts its predicted trajectory to the other ones, and we propose an obstacle prioritization scheme based on the shared trajectories to allow up-scaling of the system. 
%in terms of the number of agents.
The NMPC problem is solved using an \textit{ad hoc} solver
where PANOC is combined with an augmented Lagrangian method to compute collision-free trajectories.
%This article will present the control structure, optimization framework, as well as the distributed architecture of the multi-agent system. 
We evaluate the proposed scheme 
in several challenging laboratory experiments for up to ten aerial agents, in dense aerial swarms. 

\end{abstract}
\glsresetall %it resets the abbreviations to be redefine in the introduction
%%%%%%%%%%%%%%%%%%%%%%%%%%%%%%%%%%%%%%%%%%%%%%%%%%%%%%%%%%%%%

%%%%%%%%%%%%%%%%%%%%%%%%%%%%%%%%%%%%%%%%%%%%%%
\subsection{Introduction and Background}
%%%%%%%%%%%%%%%%%%%%%%%%%%%%%%%%%%%%%%%%%%%%%%
One of the most popular and exciting areas of robotics right now is the area of collaborative robotics, meaning that a team of robots are tasked with collaboratively performing a specified mission, which could include collaborative inspection\cite{mansouri2018cooperative}, mapping \& exploration\cite{nieto2014coordination}, search-and-rescue\cite{beck2016online} and many others. 
%Generally, and especially for the Unmanned Aerial Vehicle (UAV) case, robots are limited by battery life, and most tasks are greatly enhanced if they can be performed by teams of robots as to reduce the time to mission completion. 
The co-operation and coordination of teams of robots can be done in many different ways, but a common nomenclature is the division into centralized, decentralized, or distributed schemes. For centralized approaches all computation and planning is done via a single computational agent, which is fed all available system information. In the decentralized approach, all agents act independently only based on information available to that agent and as such allow for much greater scalability. The distributed scheme is the middle ground, where every agent computes its own decisions, but specific information is transmitted between agents that supplement the computed decisions, which allows for both great co-operation and scalability. 

When there are multiple robotic agents occupying a small space to perform a task, the collision avoidance scheme must not only avoid collisions with the environment, but also collision among agents. In this article we propose a distributed nonlinear model predictive control (DNMPC), where computed trajectories are shared among agents, and other agents in the system are considered as obstacle constraints. 
%This creates a system that has completely decoupled dynamics, but where agents are coupled only by the obstacle constraints. 
%To increase the scalability of the scheme, we propose an obstacle prioritization method that directly considers the predicted motion of every agent. 

%While Distributed Model Predictive Control (DMPC) is a very well studied subject\cite{maestre2014distributed} with many interesting applications in both industry and robotics, we shall focus on DMPC schemes specifically for UAVs.
There are many different approaches to collision avoidance schemes for multi-agent robotic systems. Rule-based schemes, such as potential functions \cite{budiyanto2015uav, mondal2017novel} or optimal control schemes \cite{hoffmann2008decentralized}, work well in low-density swarms of agents but do not include hard guarantees on safety distances or to do so must include additional restrictive rules. A differential game approach is proposed in \cite{mylvaganam2017differential} with impressive results, but relies on many conditions, and the article only offers simulation results. In \cite{mellinger2012mixed} a mixed-integer quadratic program is proposed for centralized trajectory generation, but does not allow for real-time solutions meaning it is non-reactive. Other modern solutions showing promise are the barrier functions \cite{wang2017safety}, or \textit{safety barrier certificates} which were evaluated for mini quad-copters in laboratory experiments, and \cite{verginis2020adaptive} that combined potential-like functions with adaptive control to achieve collision avoidance robust to model uncertainty.

Model predictive control (MPC) schemes for multi-agent collision avoidance  are, due to their high performance, flexibility, and ability to handle constraints, one of the most popular approaches to multi-agent collision avoidance \cite{alonso2015collision}. The ability to optimize over future predicted states allows for a direct consideration of moving obstacles, and with fast optimization algorithms allow for real-time reactive and proactive avoidance. 
MPC schemes comes in many different flavors: in terms of the system dynamics and constraints there are linear \cite{luis2020online} and nonlinear \cite{filotheou2018decentralized} solutions, and in terms of architecture there are decentralized \cite{kamel2017nonlinear}, centralized \cite{lindqvist2020collision}, and distributed \cite{dai2017distributed} schemes.

In \cite{luis2020online}, a linear UAV model is used where the decision variable of the MPC is a position-trajectory e.g. it requires a separate position-reference tracking controller to follow the computed trajectory. All agents share predicted trajectories in a distributed fashion, show great scalability, and was experimentally evaluated for up to 20 agents. 

In \cite{kamel2017nonlinear} a NMPC scheme is used for multi-agent decentralized collision avoidance. No information is shared among agents, and predicted future positions of the non-ego agents are done with a constant velocity model from initial position and velocity measurements. The motion intent of other agents is simplified, but no information transfer among agents is required. However, the experimental validation involves only two UAVs. 

In \cite{zhu2019chance} a method proposing probabilistic collision avoidance for multi-agent systems using NMPC is demonstrated, and shows experimental results for low-density and low numbers of robots, while up-scaling is only demonstrated in simulation. 

In our previous paper \cite{lindqvist2020collision} we proposed a centralized NMPC scheme, where a single computational agent optimized the trajectories of all agents. While this scheme showed good results in simulation, all centralized schemes will eventually suffer from scalability issues. %, and the direct reliance on constant communication with the central computation agent would make a field-application very limited. 

A common limitation of existing MPC approaches is that since the number of constraints (or potential-like terms) of the underlying optimization problem must remain constant during runtime, there can be situations where there are too many agents in proximity to the ego vehicle. Moreover, selecting the closest neighbors (as in \cite{luis2020online}) may disregard those --- potentially more remote --- agents that are on a direct collision course with the ego agent.

% A common limitation of existing MPC methods is that either no discussion on how to deal with situations where the number of robots in proximity exceeds the number of constraints (or potential-like costs) defined in the MPC formulation, or the use of very simplistic models such as always selecting the closest neighbors.
%
%%%%%%%%%%%%%%%%%%%%%%%%%%%%%%%%%%%%%%%%%%%%%%
\subsection{Contributions}
%%%%%%%%%%%%%%%%%%%%%%%%%%%%%%%%%%%%%%%%%%%%%%
%
%Compared to these recent State-of-the-art papers, 
We propose a distributed NMPC solution where the collision avoidance is solved for in the control layer, and as such no additional position- or velocity-tracking controller is needed. 
In our approach, each agent uses a nonlinear dynamical model and shares its predicted trajectory with all neighbouring agents 
at every sampling time. Since these trajectories are produced
by a nonlinear model, they are sufficiently close to their actual  
trajectories to allow for collision-free coordination.
%In our approach, all agents share predicted trajectories, and due to the adopted nonlinear UAV model, will better match the real behavior, or motion intent, of other agents.
Furthermore, we propose an obstacle prioritization algorithm that 
determines which agents are on the most dangerous predicted potential collision courses with
the ego agent based on their shared predicted motion. Additionally we propose an adaptive scheme for the MPC weights that facilitates challenging 
collision avoidance situations.
This allows us to sacrifice the speed of reaching the target for enforcing collision avoidance.
This leads to a highly scalable paradigm that can accommodate
large numbers of agents without compromising computational 
feasibility. 

In our proposed scheme, we use a tailored method that has been implemented in Optimization Engine (for short \texttt{OpEn}),which is an open-source code generation software for embedded nonlinear
optimization \cite{sopasakis2020open,open2019}, that is fully ROS-integrated.  
It generates Rust code, which is very fast and provably memory safe. 
\texttt{OpEn} uses PANOC (proximal averaged Newton-type method for optimal
control) \cite{stella2017simple, sathya2018embedded} combined with 
an augmented Lagrangian method \cite{sopasakis2020open,birgin2014practical} 
to account for general non-convex constraints such as the ones that 
result from collision avoidance.
To the best of our knowledge, this is the first work that offers experimental results for collision avoidance using the augmented Lagrangian method and we shall demonstrate that the proposed scheme is suitable for a fast real-time implementation.

We present results from multiple experimental scenarios 
for up to ten agents in tight formations, and with low numbers of constraints to reduce computational complexity, to demonstrate the collision avoidance capabilities of the proposed control architecture.

%%%%%%%%%%%%%%%%%%%%%%%%%%%%%%%%%%%%%%%%%%%%%%
\section{Methodology} \label{sec:methodology}
%%%%%%%%%%%%%%%%%%%%%%%%%%%%%%%%%%%%%%%%%%%%%%

%%%%%%%%%%%%%%%%%%%%%%%%%%%%%%%%%%%%%%%%%%%%%%
\subsection{System Dynamics} \label{sec:mavkinematic}
%%%%%%%%%%%%%%%%%%%%%%%%%%%%%%%%%%%%%%%%%%%%%%
The adopted system model is a nonlinear dynamic UAV model, successfully used in previous applications of NMPC for UAVs \cite{lindqvist2020collision, kamel2017model,small2019aerial}, and thus we will not go into much detail in this article. Importantly, it considers eight states namely: positions coordinates $p=[p_x,p_y,p_z]^\top$, linear velocities $v = [v_x,v_y,v_z]^\top$ as well as roll and pitch angles $\phi$ and $\theta \in[-\pi,\pi]$. $\dot{\phi}$ and $\dot{\theta}$ are modeled as first order systems to approximate the closed loop behavior of an attitude controller with inputs $\phi_{\mathrm{ref}}, \theta_{\mathrm{ref}}$, and as such the control inputs of the system are $T, \phi_{\mathrm{ref}}, \theta_{\mathrm{ref}}\in \mathbb{R}$ where $T \geq 0$ is the total mass-less thrust produced by the motors. Based on this model let us define the state vector as $x = [p, v, \phi, \theta]^\top$ and the control actions as $u=[T,\phi_{\mathrm{ref}},\theta_{\mathrm{ref}}]^\top$. 
The full dynamic model is as follows: 
\begin{subequations}
\label{eq:mavkinematic}
\begin{align}
        \dot{p}(t) &= v(t) \\ 
        \dot{v}(t) &= R(\phi,\theta) 
        \smallmat{0 \\ 0 \\ T} + 
        \smallmat{0 \\ 0 \\ -g} - 
        \smallmat{A_x & 0 & 0 \\ 0 &  A_y & 0 \\ 0 & 0 & A_z} v(t), \\ 
        \dot{\phi}(t) & = \nicefrac{1}{\tau_{\phi}} (K_\phi\phi_{\mathrm{ref}}(t)-\phi(t)), \\ 
        \dot{\theta}(t) & = \nicefrac{1}{\tau_{\theta}} (K_\theta\theta_{\mathrm{ref}}(t)-\theta(t)).
\end{align}
\end{subequations}
This model is then discretized by the forward Euler transformation to achieve the predictive form
\begin{equation}
\label{eq:prediction}
    x_{k+1} = \zeta(x_k, u_k).
\end{equation}
This model is used as the prediction model for the receding horizon NMPC-problem, where the number of predicted time instants is denoted by the prediction horizon, $N$. 
%

%%%%%%%%%%%%%%%%%%%%%%%%%%%%%%%%%%%%%%%%%%%%%%
\subsection{Control Objective} \label{sec:obj}
The main control objective of each agent is to track a given set point, 
while avoiding collisions with other agents. Additional objectives are 
not to have abruptly changing control actions and to keep the control 
actions within certain bounds. Here, we will translate these requirements
into a cost function and constraints for a nonlinear MPC formulation.

\subsubsection{Objective Function}
%%%%%%%%%%%%%%%%%%%%%%%%%%%%%%%%%%%%%%%%%%%%%%
%
Let $x_{k+j{}\mid{}k}$ and $u_{k+j{}\mid{}k}$ denote the predicted state 
and input at time $k+j$, computed at the time $k$. 
Let $\bm{x}_{k}$ and $\bm{u}_{k}$ be the vectors of all predicted states 
and inputs along the prediction horizon. 
The deviation of the predicted state from the set point, 
$x_{\mathrm{ref}}$, can be penalized using a standard quadratic function
and, likewise, we introduce a quadratic cost the input deviation.
Additionally, we introduce a penalty on successive changes in control inputs, $u_{k+j{}\mid{}k}-u_{k+j-1{}\mid{}k}$ (note that $u_{k-1{}\mid{}k} = u_{k-1}$ which is the the previous control action), and a standard quadratic terminal state cost. The overall cost is given by
\begin{multline}
\label{eq:costfunction}
J(\bm{x}_{k}, \bm{u}_{k}, u_{k-1\mid k}) = \sum_{j=0}^{N-1} \big(  \smashunderbracket{\| x_{\mathrm{ref}}-x_{k+j{}\mid{}k}\|_{Q_x}^2}{\text{State penalty}}
\\
+   \smashunderbracket{\| u_{\mathrm{ref}}-u_{k+j{}\mid{}k}\|^2_{Q_u}}{\text{Input penalty}}
+  \smashunderbracket{\| u_{k+j{}\mid{}k}-u_{k+j-1{}\mid{}k} \|^2 _{Q_{\Delta u}}}{\text{Input change penalty}}\big) \\[1.2em]
+ \smashunderbracket{\| x_{\mathrm{ref}}-x_{k+N{}\mid{}k}\|_{Q_t}^2}{\text{Terminal state penalty}},
\end{multline}
\vspace{0.35em}

\noindent where $Q_x, Q_t \in \mathbb{R}^{8\times8}, Q_u, Q_{\Delta u}\in 
\mathbb{R}^{3\times3}$ are positive definite weight matrices for the
states, terminal state, inputs and input change respectively. 
%

%%%%%%%%%%%%%%%%%%%%%%%%%%%%%%%%%%%%%%%%%%%%%%
\subsubsection{Obstacle Definition} \label{sec:const}
%%%%%%%%%%%%%%%%%%%%%%%%%%%%%%%%%%%%%%%%%%%%%%
%
We require that the ego agent does not approach any of the other agents at a 
distance closer than a safety value $r_{\mathrm{obs}}$. 
Let $p$ and $p^{\mathrm{obs}} = [p_x^{\mathrm{obs}}, p_y^{\mathrm{obs}}, 
p_z^{\mathrm{obs}}]$ denote the positions of the ego agent and a non-ego
agent respectively. Then, define
\begin{multline}\label{eq:sphericalconstraint} 
    h_{\mathrm{sphere}}(p, \xi^{\mathrm{obs}}) = (r^{\mathrm{obs}})^2 - (p_x{-} p_x^{\mathrm{obs}})^2 \\
    - (p_y{-}p_y^{\mathrm{obs}})^2 - (p_z{-}p_z^{\mathrm{obs}})^2,
\end{multline}
with $\xi^{\mathrm{obs}} = [p^{\mathrm{obs}}, r^{\mathrm{obs}}]$. The obstacle avoidance requirement is equivalent to $h_{\mathrm{sphere}} \leq 0$, that ego agent position $p$ is required to lie completely outside of the sphere defined by $\xi^{\mathrm{obs}}$. Additionally, we require that the constraint holds for predicted positions of the ego vehicle, $p_{k + j\mid k }$, and predicted obstacle positions $p^{\mathrm{obs}}_{k + j \mid k}$ along the prediction horizon.

\subsubsection{Input bounds} \label{sec:input_constraints}
%%%%%%%%%%%%%%%%%%%%%%%%%%%%%%%%%%%%%%%%%%%%%%
For a UAV system based on \eqref{eq:mavkinematic}, an attitude controller will only be able to stabilize the UAV within a specific range of $\phi$ and $\theta$, therefore the control actions commanded by the NMPC and especially $\phi_{\mathrm{ref}}$ and  $\theta_{\mathrm{ref}}$, should be constrained. Additionally, to further limit the acceleration of the UAV, we also place such limits on $T$. For this purpose we impose the bounds
\begin{equation}
\label{eq:input_const}
u_{\min} \leq u_{k+j\mid k} \leq u_{\max}.
\end{equation}

\subsection{Obstacle Prioritization}\label{sec:prio}
%An important concept in multi-agent systems is scalability. For a system such as the one described the dynamics of each agent is decoupled from the others, and the agents are coupled only by the constraints. If computational efficiency and the speed and accuracy of optimization algorithms were infinite each agent would always be coupled to every other agent by collision avoidance constraints ($N_a = N_{obs} + 1$). Unfortunately this is not the case, and as such to allow scalability in terms of the number of agents in the distributed scheme there needs to be some level of obstacle prioritization. This could be done by assuming a larger radius of influence around each agent, where the ego agent would consider all agents within this volume as obstacles, or sort them by distance to the ego agent, but this is a too simplified structure. Assuming that each agent has access to the predicted trajectories of nearby agents, we can formulate a way to prioritize which agents the ego agent should consider as dynamic obstacles based on the shared trajectories, e.g. the motion intent of other agents, and if they intend to move close to the ego agent at future time steps. Based on this goal, we propose the Algorithm 1 as the obstacle prioritization algorithm, which can be described as follows: 

An important concept in multi-agent system is scalability. Assuming a system composed of $N_a$ agents, ideally each agent should be able to form obstacle constraints with all other agents, such that the number of obstacles $N_{\mathrm{obs}} = N_a - 1$. Due to limitation in computation power and the speed of optimization algorithms, this is not always possible for high numbers of agents. Instead, we need to choose $N_{\mathrm{obs}} < N_a - 1$ necessitating some kind of obstacle prioritization where the ego agent takes into account a limited number of $N_{\mathrm{obs}}$ other agents. Assuming access to the predicted trajectories, $\bm{u}_{k-1}^{\mathrm{obs},i}$ and measured states $\hat{x}^{\mathrm{obs},i}_k$  of nearby agents, we want the prioritization scheme to be fully based on the motion intentions of each agent. For this purpose we propose Algorithm \ref{alg:obstable_prioritization} as the obstacle prioritization scheme, which can be described as follows:
\begin{itemize}
    \item Using the NMPC prediction model \eqref{eq:prediction} we can describe any $x_{k-1 + j \mid k - 1}$ from $\bm{u}_{k-1}$ and $\hat{x}_k$,  and similarly using the shared NMPC solutions $\bm{u}_{k-1}^{\mathrm{obs},i}$ and $\hat{x}^{\mathrm{obs},i}_k$ to describe $x^{\mathrm{obs},i}_{k-1+j \mid k - 1}$ for $i = 1 \ldots N_a - 1$.
    \item Calculate the predicted Euclidean distances between the ego agent and all other agents at the current and future time instants $j=0,\ldots, N$
    \item An agent is prioritized if the distance is below a threshold specified by $r^{\mathrm{obs},i} + d_s$ where $d_s$ is a positive safety distance.
    This is done via the following weighted sum as a gauge for the prioritization 
    \[
      w_i = \sum_{j=0}^{N}\alpha(d_{i,j}, v^{\mathrm{obs},i}_{k-1 + j \mid k - 1}) \beta(j),
    \] 
    where $d_{i,j}$ denotes the distance to the $i$-\textit{th}
    agent at time instant $j$. 
    Let $\alpha(d_{i,j}$, $v^{\mathrm{obs},i}_{k-1 + j \mid k - 1})$
    and $\beta(j)$ be decreasing functions in $d_{i,j}$ and $j$ 
    respectively, and as such the scheme prioritizes agents that are at a 
    closer predicted distance, and at less distant predictions.
    
    \item We also add an extra safety protocol by adding a large number $M$ to $w_i$ if the agents are closer than the obstacle radius $r^{\mathrm{obs},i}$ at the current measured positions ($j = 0$) to always prioritize agents that directly violate the obstacle constraint at the current time instant.
    
    \item Obstacle trajectories $(\xi_{j}^{{\rm obs}, i})_{i, j}$ are then sorted by the corresponding values in $\bm{w}$ in descending order, to prioritize the $N_{\mathrm{obs}}$ trajectories that produced the largest sums $w_i$.
\end{itemize}
For the weight functions we propose simple expressions with the desired functionality, and in this article we are using $\alpha(d_,v) = (1 -\frac{d}{r^{obs} + d_s})^2 \| v \|$ (velocity compensation since faster moving obstacles spend fewer time instants in $r^{\mathrm{obs},i} + d_s$) and $\beta(j) = \frac{N}{(j+1)^a}$ with $a$ being a tuning constant to describe the relative emphasis on closer versus more distant time instants.

\begin{algorithm}[ht!] 
\SetAlgoLined
\textbf{Inputs:} $\hat{x}_k, \bm{u}_{k-1}, \hat{x}^{\mathrm{obs},i}_k, \bm{u}_{k-1}^{\mathrm{obs},i}, N_a, N_{\mathrm{obs}}, \bm{r}^{\mathrm{obs}}, d_s$  \\
\KwResult{From the shared trajectories and measured states decide which $N_{\mathrm{obs}}$ agents should be considered as obstacles}
 \For{$i = 1,N_a-1$}{
     \For{$j = 0,N$}{
         Compute $p_{k-1+j \mid k - 1}$, $p^{\mathrm{obs},i}_{k-1 + j \mid k - 1}$ and $v^{\mathrm{obs},i}_{k-1 + j \mid k - 1}$\\
         $d \gets \| p_{k-1 + j\mid{}k-1} - p^{\mathrm{obs},i}_{k-1+j \mid k - 1} \|$\\
         $v_\mathrm{m} \gets \| v^{\mathrm{obs},i}_{k-1 + j \mid k - 1} \|$ \\
        \uIf{$(d \leq r^{\mathrm{obs},i})$ and $(j = 0)$}{
        $w_i \gets w_i + M$
       }
       \ElseIf{$d \leq r^{\mathrm{obs},i} + d_s$}{
       $w_i \gets w_i + (1 - \frac{d}{r^{\mathrm{obs}} + d_s})^2 v_\mathrm{m} \frac{N}{(j+1)^a}$
       }
    }
}
Sort in descending order: $(\xi_j^{{\rm obs}, i})_{i, j}$ by corresponding element in $w_i$\\
$(\xi_{\mathrm{prio},j}^{{\rm obs}, i})_{i, j}  \gets [(\xi_j^{{\rm obs}, 1})_{j}, (\xi_j^{{\rm obs}, 2})_{j}, \ldots, (\xi_j^{{\rm obs}, N_{\mathrm{obs}}})_{j}]$\\
\textbf{Output:} $(\xi_{\mathrm{prio},j}^{{\rm obs}, i})_{i, j}$  \\
 \caption{Obstacle Prioritization}\label{alg:obstable_prioritization}
\end{algorithm}
%%%%%%%%%%%%%%%%%%%%%%%%%%%%%%%%%theta%%%%%%%%%%%%%
\subsection{Embedded Model Predictive Control} \label{sec:solver}

\subsubsection{NMPC Problem}
In light of the aforementioned objectives and constraints, 
the obstacle avoidance problem for each agent leads to the following
constrained nonlinear optimal control problem
\begin{subequations}\label{eq:nmpc}
\begin{align}
    \operatorname*{Minimize}_{
        \bm{u}_k, \bm{x}_k
    } \,
    & J(\bm{x}_{k}, \bm{u}_{k}, u_{k-1\mid k})
    \\
    \text{subj. to:}\,& 
    x_{k+j+1\mid k} {=} \zeta(x_{k+j\mid k}, u_{k+j\mid k}), j{\in}\N_{[0, N-1]},
    \\
    &u_{\min} \leq u_{k+j\mid k} \leq u_{\max}, j\in\N_{[0, N-1]},
\label{eq:nmpc:input_constraints}
    \\
    &h^i_{\mathrm{sphere}}(p_{k+j\mid k}, \xi^{\mathrm{obs},i}_{\mathrm{prio},j}) \leq 0,\,
     j\in\N_{[0, N]},\\
     &\ i{\in}\N_{[1, N_{\mathrm{obs}}]},\\
     &x_{k\mid k} {}={} \hat{x}_k,
\end{align}
\end{subequations}
where $\hat{x}_k$ is the current estimated system state.
This problem needs to be solved at every sampling time by each agent taking into 
account the trajectories, $(\xi_j^{{\rm obs}, i})_{i, j}$, that the other agents 
have shared. The problem yields an optimal sequence of control actions, the first
one of which is applied to the ego vehicle and the optimal predicted trajectory
is broadcast to all surrounding agents.
% e.g. in terms of the spherical constraint expression the set of points outside of the sphere. Meaning, we want to minimize the objective function $J(\bm{x}_{k}, \bm{u}_{k}, u_{k-1\mid k})$ based on \eqref{eq:costfunction}, while being subject to the discrete system model $\zeta(x_{k+j\mid k}, u_{k+j\mid k})$, the input constraints defined by \eqref{eq:input_const}, and the $N_{obs}$ spherical obstacle constraints described in \ref{sec:const}. As such, this NMPC problem encapsulates the desired control objectives. 

\subsubsection{Embedded numerical optimization}
The optimization problem in Equation \eqref{eq:nmpc} can be 
written concisely in the following form
\begin{subequations}\label{eq:open_problem}
\begin{align}
    \mathbb{P}(x_{k\mid k})
    {}:{}
    \operatorname*{Minimize}_{\bm{u}_k \in U}\,& f(\bm{u}_k; x_{k\mid k})\label{eq:nmpc:1}
    \\
    \text{subject to:}\,& F(\bm{u}_k; x_{k\mid k}) \leq 0,\label{eq:nmpc:2}
\end{align}
\end{subequations}
where $f({}\cdot{}; x_{k\mid k}):\R^{3N}\to\R$ is a Lipschitz-differentiable function and 
$F({}\cdot{}; x_{k\mid k}):\R^{3N}\to\R^{NN_{\rm obs}}$ 
is a differentiable mapping with Lipschitz-continuous Jacobian. 
Here the optimization is carried out over sequences of control actions, 
$\bm{u}_k$, while the sequence of states, $\bm{x}_k$, has been eliminated
following what is known as the \textit{sequential} or \textit{single shooting}
formulation \cite{small2019aerial}.
The cost function in \eqref{eq:nmpc:1} is defined by \eqref{eq:costfunction},
where the state sequence has been eliminated and the constraints in 
Equation \eqref{eq:nmpc:2} correspond to the obstacle avoidance constraints
discussed in Section \ref{sec:const}.
The sequence of control actions, $\bm{u}_k$, is constrained in a set $U\subseteq\R^{3N}$,
which is the rectangle defined by the constraints of Equation \eqref{eq:input_const}.

Problem $\mathbb{P}(x_{k\mid k})$ in \eqref{eq:open_problem} is solved by using 
the augmented Lagrangian method. The associated augmented Lagrangian function is 
\begin{multline}
    L_c(\bm{u}_k, \bm{v}_k, \bm{y}_k; x_{k\mid k}) 
    = 
    f(\bm{u}_k; x_{k\mid k})\\
    +
    \bm{y}_k^\top \left(F(\bm{u}_k; x_{k\mid k}) - \bm{v}_k\right)
    +
    \tfrac{c}{2}\|F(\bm{u}_k; x_{k\mid k}) - \bm{v}_k\|^2,
\end{multline}
defined for $\bm{v}_k{}\in{}U$ and $\bm{v}_k{}\leq{}0$
It has been shown in \cite{sopasakis2020open} that
\begin{multline}
    \min_{\bm{u}_k{}\in{}U, \bm{v}_k{}\leq{}0}
    L_c(\bm{u}_k, \bm{v}_k, \bm{y}_k; x_{k\mid k}) 
    =-\tfrac{1}{2c}\|\bm{y}_k\|^2 \\
    + \min_{\bm{u}_k{}\in{}U} \psi(\bm{u}_k{}; c, \bm{y}_k, x_{k\mid k}),
\end{multline}
where $\psi$ is defined as 
\begin{multline*}
    \psi(\bm{u}_k{}; c, \bm{y}_k, x_{k\mid k})
    {}={}
    f(\bm{u}_k; x_{k\mid k})
    \\
    +
    \tfrac{c}{2}
    \left\| 
      F(\bm{u}_k; x_{k\mid k}) + \tfrac{1}{c}\bm{y}_k
      -
      [F(\bm{u}_k; x_{k\mid k}) + \tfrac{1}{c}\bm{y}_k]_{-}
    \right\|^2,
\end{multline*}
where $[z]_{-} = \max\{0, -z\}$.
Function $\psi$ is continuously differentiable and has a Lipschitz-continuous
gradient, therefore the ``inner'' optimization problem 
$\min_{\bm{u}_k{}\in{}U} \psi$ can be solved very efficiently using PANOC. The gradient of $\psi$ can be determined by means
of automatic differentiation.
This leads to Algorithm \ref{alg:alm}.

\begin{algorithm}[ht!] 
\SetAlgoLined
\textbf{Inputs:} \(\bm{u}_k^0\in\R^{3N}\) (initial guess), 
           \(x_{k\mid k}\in\R^{8}\) (parameter),
           \(\bm{y}_k^0\in\R^{NN_{\rm obs}}\) (initial guess for the Lagrange multipliers), 
           \(\epsilon, \delta > 0\) (tolerances),
           \(\rho\) (penalty update coefficient), 
           \(c_0\) (initial penalty),
           \(\theta\) (sufficient decrease coefficient),
          \(M \gg 1\) (large constant) 
           \\
\KwResult{\((\epsilon, \delta)\)-approximate solution $(\bm{u}_k^\star, \bm{y}_k^\star)$}
 \For{$\nu=0,\ldots, \nu_{\mathrm{max}}$}{
  $\bar{\bm{y}}_k^{\nu}{}={} \max\{0, \min\{M, \bm{y}_k^\nu\} \}$
  \\
  Compute a solution
  \(
  \bm{u}^{\nu+1} \in \argmin_{\bm{u}_k{}\in{}U} \psi(\bm{u}_k{}; c_\nu, \bar{\bm{y}}_k^\nu, x_{k\mid k})
  \)
  with tolerance $\bar\epsilon$ and initial guess $\bm{u}^\nu$ using PANOC
  \\
  Update $\bm{y}_k^{\nu+1}$ by
  \begin{multline}
      \bm{y}_k^{\nu+1} {}={} \bar{\bm{y}}_k^{\nu+1} + c_\nu(F(\bm{u}_k^{\nu+1}; x_{k\mid k}) 
      \\
      - \left[ F(\bm{u}^{\nu+1}; x_{k\mid k}) + c^{-1}_\nu\bar{y}^\nu) \right]_{-}
  \end{multline}
  \\ 
  $z_{\nu+1} {}={} \| \bm{y}_k^{\nu+1} - \bm{y}_k^\nu \|_\infty$
  \\
  \uIf{$z_{\nu+1} \leq c_\nu \delta$ and $\bar{\epsilon}_\nu \leq \epsilon$}{
      \textbf{return} $(\bm{u}_k^\star, \bm{y}_k^\star)=(\bm{u}_k^{\nu+1}, \bm{y}_k^{\nu+1})$
      \\
  }
  \ElseIf{$\nu>0$, $z_{\nu+1} > \theta z_{\nu}$}%
         {$c_{\nu+1} = \rho{}c_{\nu}$}
  $\bar\epsilon_{\nu+1} = \tfrac{1}{2}\bar{\epsilon}_{\nu}$
  } % -------------  end for loo

 \caption{Augmented Lagrangian method for solving $\mathbb{P}(x_{k{}\mid{}k})$}
 \label{alg:alm}
\end{algorithm}

The most important tuning parameter of the algorithm is the penalty 
update parameter $\rho > 1$. Typical values that seem to work well 
are between $1.1$ and $5$. A value close to $1$ will make increase the 
penalty parameter slowly, therefore, the optimal solution $\bm{u}_k^\nu$
at iteration $\nu$ will be a good initial guess for the inner optimization 
problems, which are expected to converge faster, but this will come at the 
expense of more ALM iterations. On the other hand, a larger value of $\rho$
will lead to a lower number of ALM iterations, but the initial guess 
for the inner problems will not be as good, this requiring a higher 
computational effort. 

Overall, the algorithm has low
memory requirements, involves only simple operations (no linear 
systems or factorisations) and is very fast. Upon termination returns an $\epsilon$-suboptimal
and $\delta$-infeasible solution.

The algorithm is implemented in \texttt{OpEn} which generates Rust code that solves the parametric problems $\mathbb{P}(x_{k{}\mid{}k})$ for each agent \cite{sopasakis2020open,open2019}.

\subsection{Adaptive NMPC Weights}
The optimal Lagrange multipliers, $\bm{y}_k^\star$, can be thought 
as indicators of how much the optimal trajectories need to ``bend'' 
to avoid the obstacles.
The idea is to use $\bm{y}_k^\star$ to update the reference tracking
weights so that obstacle avoidance is prioritized over set point 
tracking.

Let $Q_p$ define the first three diagonal elements of 
positive-definite weight matrix $Q_x$. 
% The adaptive scheme should also consider at which time step $j$ that $\bm{y}_k^\star$ is non-zero, and place a greater emphasis closer to the current time step.  
We introduce a scaling factor that adapts $Q_p$ from $Q_{p,\min}$ to $Q_{p,\max}$ based on Lagrange multiplier $\bm{y}_k^\star$ as follows:
\begin{equation}\label{eq:adaptive}
    Q_{p} = Q_{p,\min} + \frac{ Q_{p,\max} - Q_{p,\min}}{\sum_{l=0}^{N_{\rm obs}N} W_l y^*_{k,l} + 1}
\end{equation}
where $W_l$ is some weight that is decreasing with respect to elements in $y^*_{k,l}$ that represents constraints at more distant future time instants, which in our formulation results in
$W_l = b(1 - \frac{l\mod N}{N})$ where $b$ is a tuning constant. This heuristic seems to work well in practise. All elements in $Q_p$ are scaled by the same factor, but it is enough for reducing the general emphasis on reference tracking. Note that the terminal state penalty still promotes the UAV to be as close to its end goal as possible but only at $j = N$.

%%%%%%%%%%%%%%%%%%%%%%%%%%%%%%%%%%%%%%%%%%%%%%
\section{Results} \label{sec:results}
\subsection{Experiment Set-up and NMPC Tuning} \label{sec:setup}
The platform used for the experiments is the Crazyflie Nano 2.0 \cite{bitcraze}, using a ros-stack developed for the Crazyflie \cite{crazyflieROS} for communication with the agents and low-level attitude control. These are small lightweight platforms that do not carry their own computational devices (which does imply some communicated delays), but are great for showing scalability. As such, computation for all agents is done on a single Lenovo ThinkPad T490, with a 1.8GHz Core i7 CPU, with controllers for each UAV running as separate ROS-nodes \cite{quigley2009ros}. All state measurements $\hat{x}= [\hat{p}, \hat{v}, \hat{\phi}, \hat{\theta}]^\top$ are made by a Vicon motion-capture system and a 3-point median filter with outlier rejection to estimate velocities from Vicon position measurements.\\ 
The proposed control structure has a large number of model and tuning parameters; model parameters are $A_x = 0.1, A_y = 0.1, A_z = 0.2$, $K_\theta, K_\phi = 1$ and  $T_\phi, T_\theta = 0.5$. For the objective function weight matrices are chosen as $Q_x = \operatorname{diag}(Q_{p,1}, Q_{p,2}, Q_{p,3}, 6, 6, 6, 8,8)$ with, according to \eqref{eq:adaptive}, $Q_{p,min} = \operatorname{diag}(1,1,15)$ and $Q_{p,max} = \operatorname{diag}(6,6,45)$, $Q_u = \operatorname{diag}(5,10,10)$, $Q_{\Delta u} = \operatorname{diag}(10, 20,20)$ and lastly $Q_t = \operatorname{diag}(40, 40, 150, 20, 20, 30, 30)$. The NMPC prediction horizon is set to $N = 40$ with a sampling time of $50ms$, which implies a prediction of two seconds. Adaptive weight tuning constant $b$ is set to $0.01$, and j-scaling constant $a$ in the prioritization scheme is set to $0.7$, and $d_s =$ \unit[0.2]{m}.

The NMPC constraints described by \eqref{eq:input_const} are $u_{min} = [5, -0.25, -0.25]$ and $u_{max} = [12.5, 0.25, 0.25]$, while obstacle radii are $r^{obs} =$ \unit[0.4]{m}. Note that small inaccuracies in localization, model mismatch, inherent delays in the system and solver tolerances will always prevent this distance from being perfectly maintained. This should be compensated by increasing the obstacle radii above the safety-critical condition which for our choice of platform is around \unit[0.3]{m}. 

Lastly, the number of parametric spherical obstacles in the NMPC formulation is set to $N_{\mathrm{obs}} = 3$, while the penalty update coefficient $\rho = 1.5$ and initial penalty $c_0 = 1000$. 

\subsection{Experimental Validation Results}
\begin{figure}[ht!]
\centering
\setlength{\belowcaptionskip}{-10pt}
  \includegraphics[width=\linewidth]{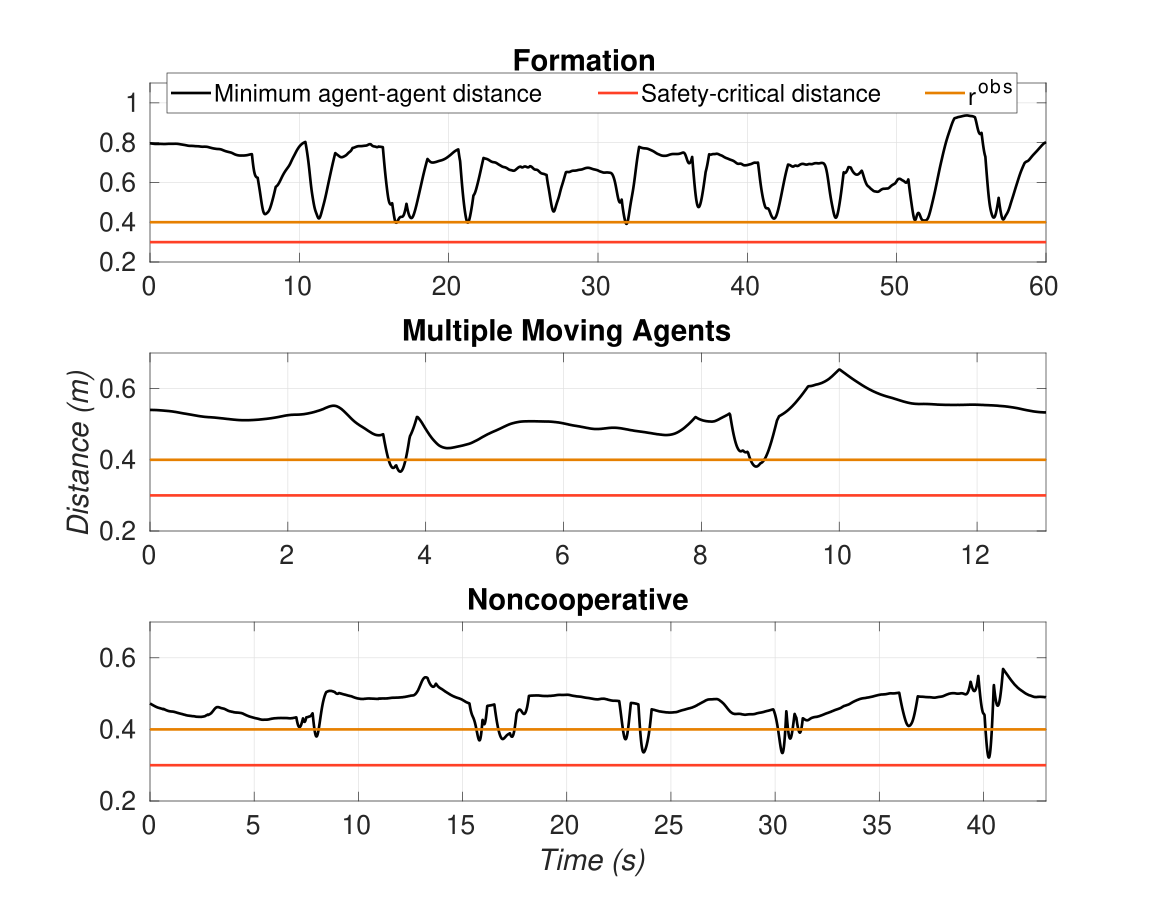}
  \caption{Minimum agent-agent distances throughout the three experiments}
  \label{fig:dist}
\end{figure}

\begin{figure}[ht!]
\centering
\setlength{\belowcaptionskip}{-10pt}
  \includegraphics[width=\linewidth]{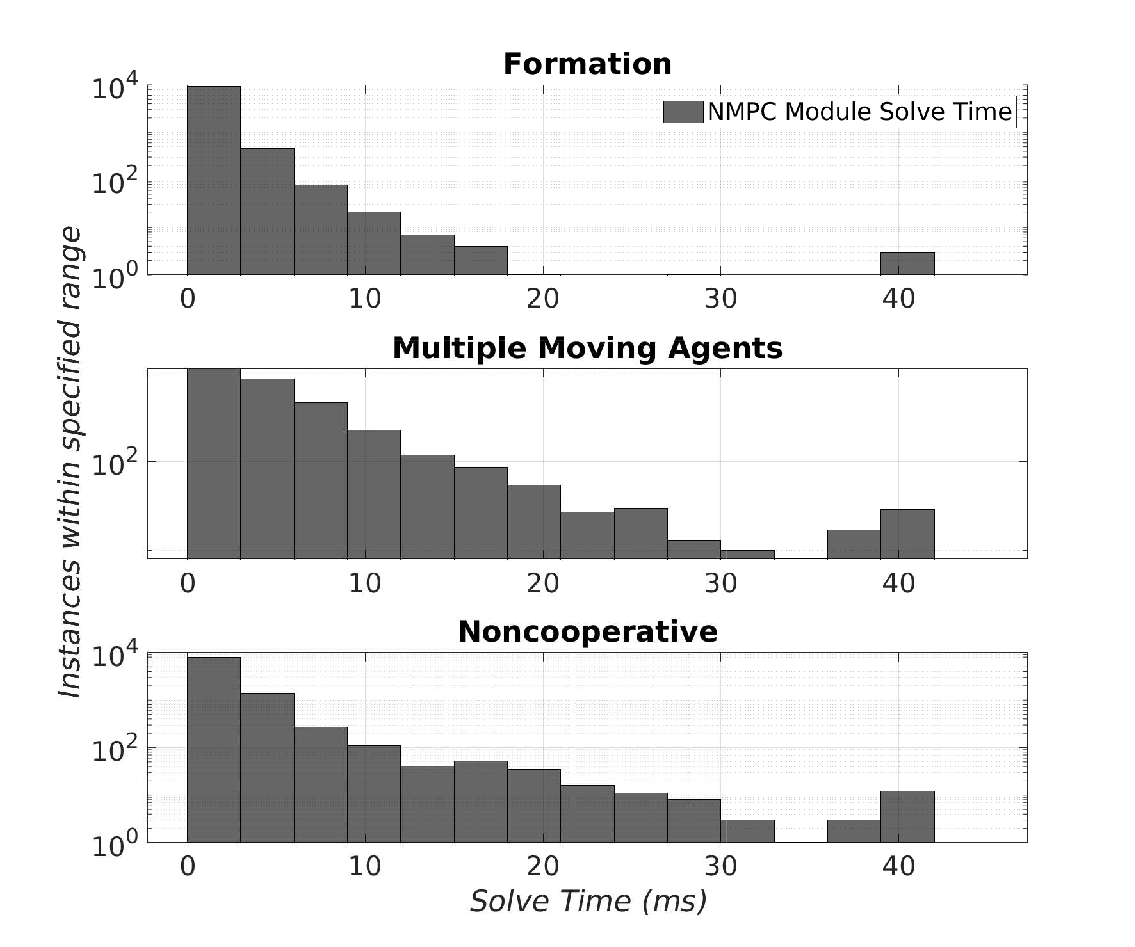}
  \caption{Histograms of solver times of the NMPC module}
  \label{fig:solvertime}
\end{figure}

To evaluate the proposed distributed collision avoidance method, we present a series of challenging laboratory experiments in which a team of UAVs performs individual position reference tracking in tights formations and challenging collision avoidance scenarios. We include up to ten agents and in the final experiment we also present the addition of a non-cooperative UAV agent to the system. Figure \ref{fig:dist} shows the minimum agent-agent distances during all three experiments, while Figure \ref{fig:solvertime} includes histograms of solver times of the NMPC module. 
For visualizing the experiments and seeing the real-time behavior of the agents we strongly suggest the reader to watch the experiment video found at \url{https://youtu.be/3kyiL6MZaag}.
\subsubsection{Position swapping in tight formations}
\begin{figure}[ht]
\centering
\setlength{\belowcaptionskip}{-10pt}
  \includegraphics[width=\linewidth]{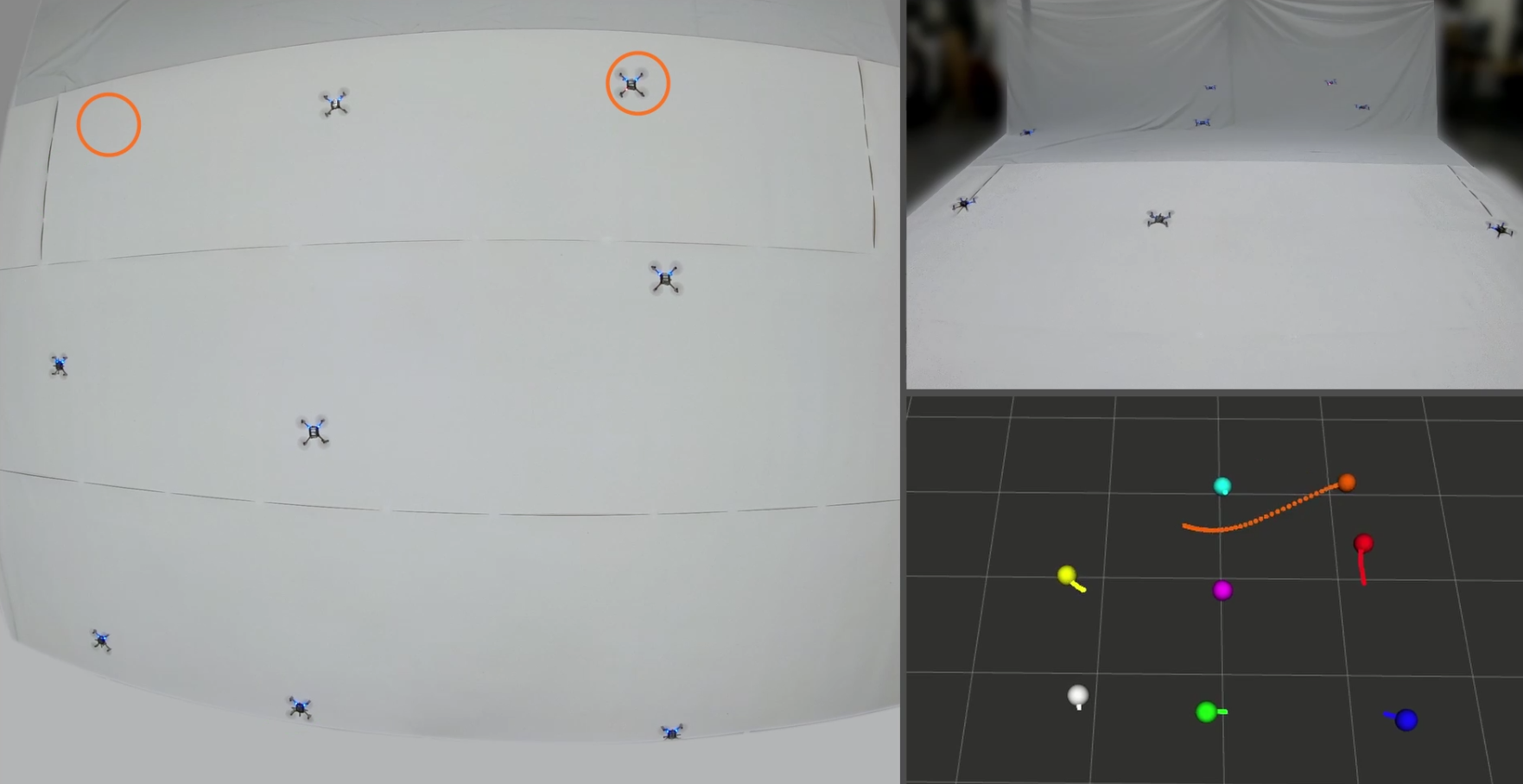}
  \caption{Experiment set-up for first experiment. Top view, side view and visualization of predicted (position) trajectories in rviz. Agents are one-by-one tasked to move positions while the formation is maintained.}
  \label{fig:exp1}
\end{figure}

Figure \ref{fig:exp1} shows the experiment set-up. The task is for the agents to hold the formation while every five seconds for one minute one agent moves to fill the unoccupied spot. The obstacle prioritization scheme allows the moving agent to quickly couple with the relevant agents as to move through the formation without collisions. Additionally, due to the sharing of the motion intentions directly though the NMPC trajectory, the formation can (mostly) be maintained as other agents see that no collision is imminent. 

The minimum agent-agent distance during the experiment was \unit[0.38]{m}. Due to the complexity and large horizon of the NMPC problem, the optimizer rarely (0.03\% of instances) did not converge to the specified tolerance within the bounds for solver time, set at \unit[40]{ms} to never have run-time issues, and the non-converged solution is instead applied to the system. The average solver time of all agents through the experiment was \unit[1.46]{ms}.

\subsubsection{Avoidance with multiple simultaneously moving agents}
\begin{figure}[ht]
\centering
\setlength{\belowcaptionskip}{-10pt}
  \includegraphics[width=\linewidth]{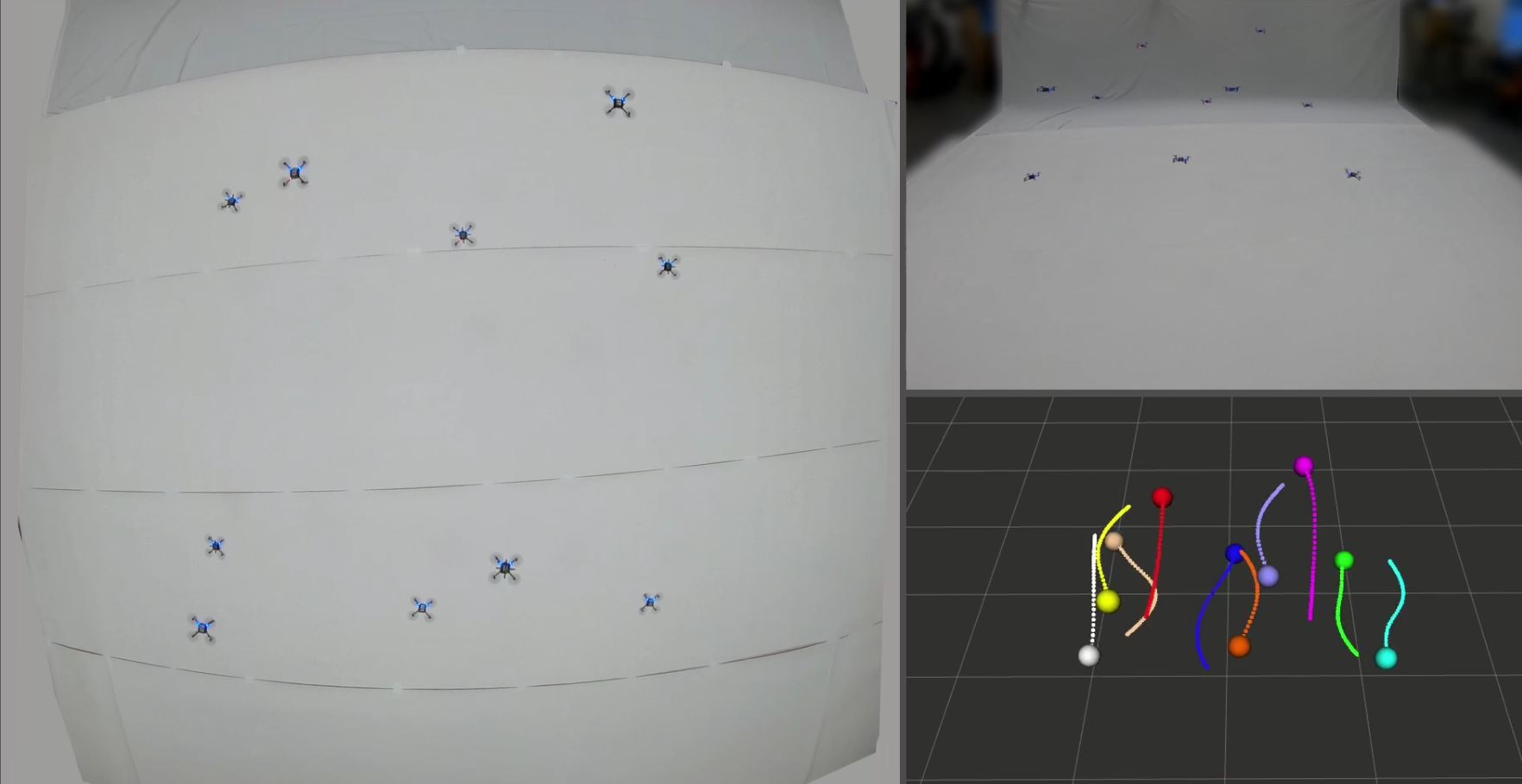}
  \caption{Experiment set-up for second experiment.  Two teams of five agents swap positions all at once.}
  \label{fig:exp2}
\end{figure}

A more challenging scenario in terms of collision avoidance is when there are multiple moving agents on direct collision courses. Figure \ref{fig:exp2} shows the set-up, where two teams of five agents are tasked to swap positions with the opposing team while maintaining collision-free paths, repeated two times. This set-up challenges the obstacle prioritization, as the closest agents are not the ones on a collision course. 
%If we only proposed to consider the closest-by-distance agents as obstacles for the ego-agent the obstacle avoidance maneuvers could not be initialed in time as to avoid collisions with the more-distant-but-on-a-collision-course agents on the opposite side. 

The agents successfully swap positions with a minimum distance of \unit[0.37]{m}. Figure \ref{fig:solvertime} shows a higher average solver time due to the shorter but more intensive experiment. The average solver time was \unit[5.35]{ms}, with around 0.7\% of instances reaching the maximum allowed solver time of \unit[40]{ms}. 
%It should be noted that when the optimizer does not converge, the optimization problem is still re-solved at 20Hz and a new optimal solution is found at the following time step, without a noticeable decrease in the maintained safety distance from the previous experiment. 

\begin{figure}[ht!]
\centering
\setlength{\belowcaptionskip}{-10pt}
  \includegraphics[width=\linewidth]{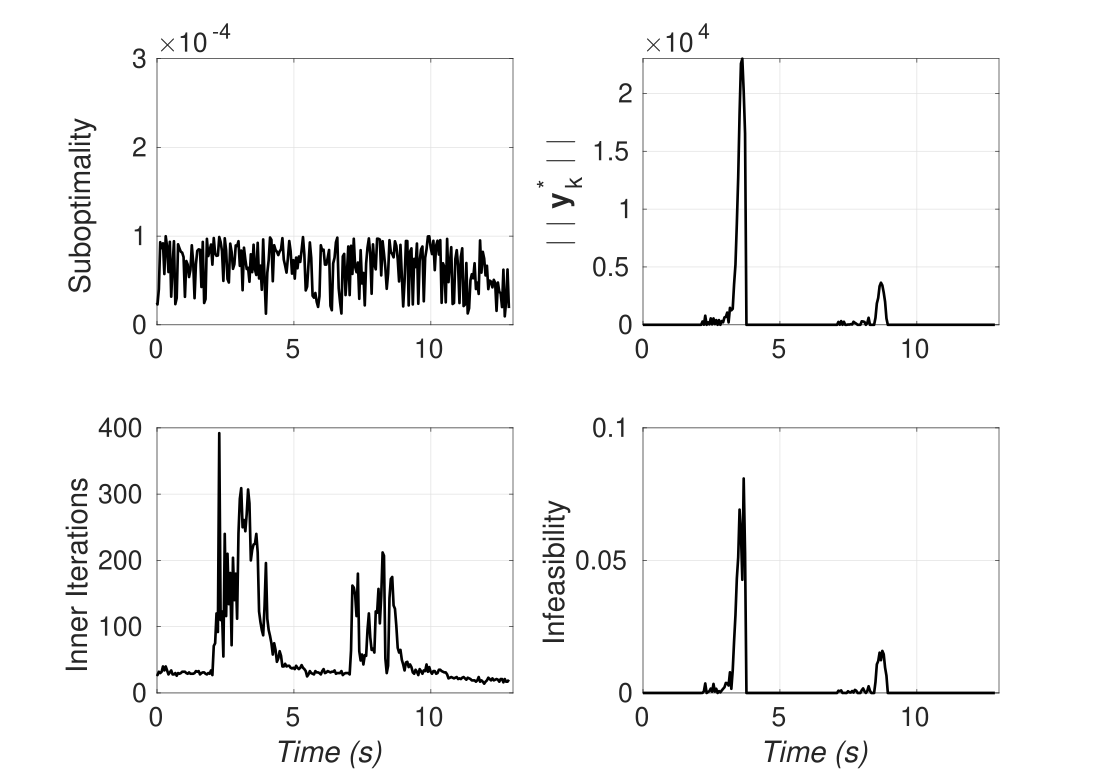}
  \caption{Solver data from one of the agents. The suboptimality described by the norm of the fixed-point residual (top left), the norm of the Lagrange Multipliers (top right), Inner iterations (bottom left) and the constraint infeasibility (bottom right).}
  \label{fig:solverinfo}
\end{figure}

Figure \ref{fig:solverinfo} displays solver data for one of the agents during the experiment: inner iterations, the suboptimality described by the norm of the fixed-point residual seen as a measure for the quality of the solution, the norm of the vector of Lagrange multipliers ,$\|\bm{y}^*_k\|$, as well as the $\delta$-infeasibility. The suboptimalty is kept at the specified solver tolerance of $10^{-4}$. As is expected the Lagrange Multipliers and $\delta$-infeasibility see a spike as the avoidance maneuvers are initiated and then rapidly drop back to zero. 

%Comparing to the distance plot in Figure \ref{fig:dist} we can see that the other solver parameters are closely tied to the challenging avoidance maneuvers, where more iterations are required to solve the constrained problem and similarly the Lagrange Multipliers quickly increase at the start of the movement but quickly drop in magnitude once the avoidance maneuver is complete. 

\subsubsection{Introducing a non-cooperative obstacle to the system}
While the distributed scheme works very well for maintaining the desired agent-agent distance, let us investigate the addition of a non-cooperative obstacle in the form of a manually controlled UAV. We use the common \cite{kamel2017nonlinear, lindqvist2020nonlinear} constant velocity model ($\dot{p}^{\mathrm{obs,nc}} = v^{\mathrm{obs,nc}}$) for moving obstacles. Discretizing the linear model we can predict obstacle positions $p^{\mathrm{obs,nc}}_{k + j \mid k}$ from the measured state $\hat{x}^{\mathrm{obs,nc}} = [\hat{p}^{\mathrm{obs,nc}}, \hat{v}^{obs,nc}]$ and add it to the list of predicted obstacle trajectories that are evaluated in the obstacle prioritization scheme. 

\begin{figure}[ht]
\centering
\setlength{\belowcaptionskip}{-10pt}
  \includegraphics[width=\linewidth]{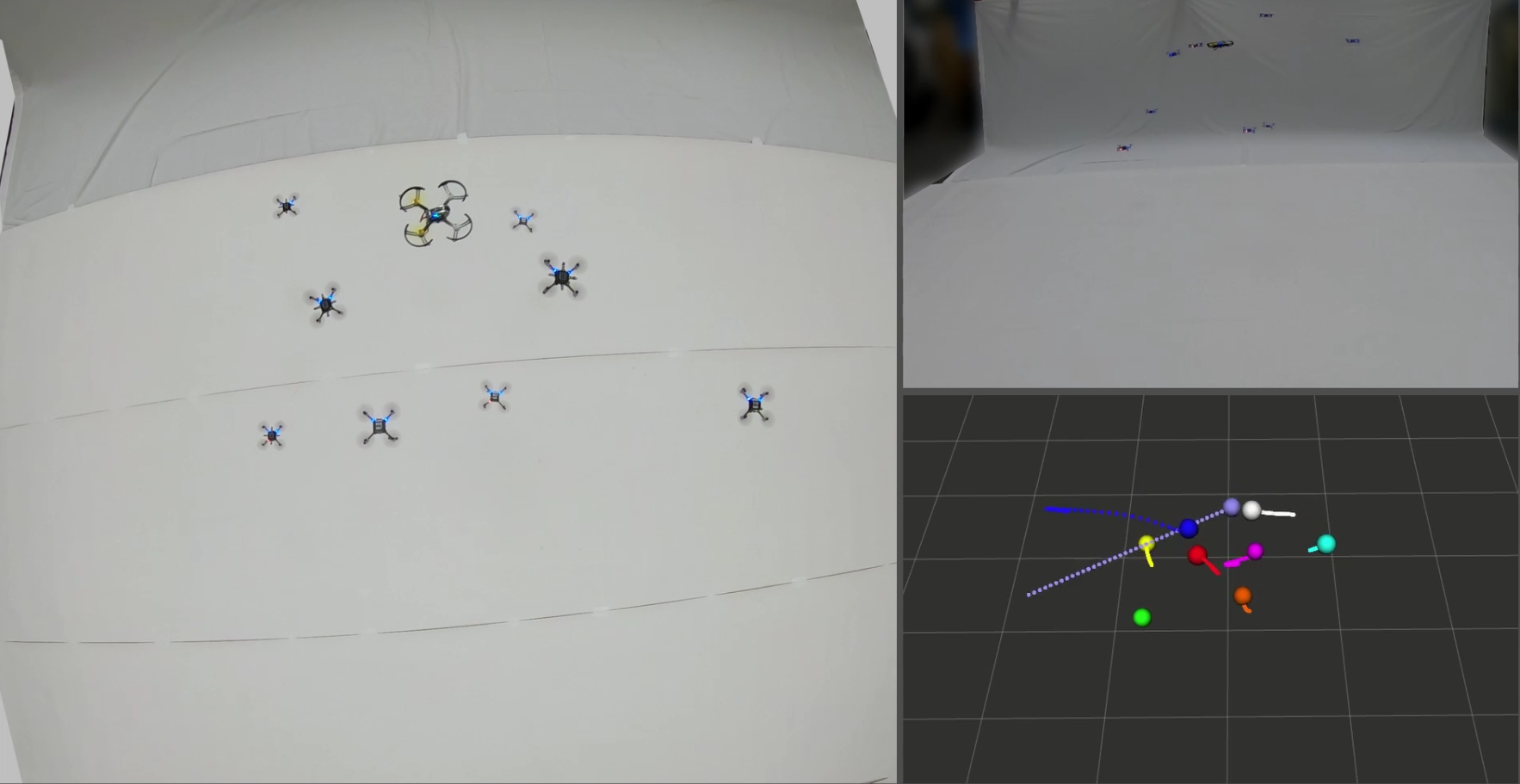}
  \caption{Experiment set-up for third experiment. The larger UAV (purple) is manually controlled and act as the non-cooperative obstacle, here seen disrupting multiple agents.}
  \label{fig:exp3}
\end{figure}

The task is to maintain a tight 8-agent formation, while a manually controlled UAV is flown through and generally tries to disrupt the formation, shown in Figure \ref{fig:exp3}. All agents must keep the required distance to other agents in the distributed scheme, while also avoiding the non-cooperative obstacle. 
The minimum distances in this experiment had an absolute minimum value of \unit[0.33]{m} and a slightly lower over-all performance. Individual agents correctly select avoidance maneuvers that avoid the non-cooperative obstacle while minimally disrupting other agents. The simplified and non-dynamic prediction model of the non-cooperative agent cannot predict sudden turns or irregular movements, but despite this the safety-critical distance is still maintained and there was no collisions with any agent. The average solver time was \unit[2.49]{ms}, with a 0.12\% non-convergence rate.

%%%%%%%%%%%%%%%%%%%%%%%%%%%%%%%%%%%%%%%%%%%%%%%%%%%%%%%%%%%%
\section{Conclusions} \label{sec:conclusion}
%%%%%%%%%%%%%%%%%%%%%%%%%%%%%%%%%%%%%%%%%%%%%%%%%%%%%%%%%%%%
In this paper we have presented a novel distributed collision avoidance scheme for Unmanned Aerial Vehicles (UAVs).
%, based on Distributed Nonlinear Model Predictive Control (DNMPC). To reduce computational complexity and allow the scheme to be greatly scaled up, we also proposed an obstacle prioritization scheme based on the shared NMPC trajectories. The scheme was evaluated in multiple challenging laboratory experiments, in tight formations, and with added non-cooperative obstacles. 
Through the presented results, we can conclude that the NMPC scheme based on the Optimization Engine (\texttt{OpEn}) can provide real-time collision-free trajectories for all agents using a prediction horizon of two seconds. Also, the implemented augmented Lagrangian method has been demonstrated to be applicable to collision-avoidance constraints with tight run-time requirements.
The obstacle prioritization correctly assigns which agents are to be considered as obstacles in time to safely avoid collisions, and allows a NMPC formulation with a low set number of constraints to fluidly deal with a large number of dynamic obstacles.
%provides an alternative answer to the question of when it is no longer appropriate to solve for collision avoidance in the control layer due to problem complexity\cite{lindqvist2020nonlinear}. 
It has also been shown that a constraint based obstacle avoidance scheme akin to the one we propose is much improved by good obstacle-trajectory information, as seen by the decreased performance when introducing the non-cooperative obstacle with a simplified predicted trajectory, a result similarly observed in \cite{zhu2019chance}. NMPC  lends itself to a distributed formulation since the motion intentions of all agents at every time instant is known based on the NMPC trajectories, but the question of how to accurately predict future obstacle positions of non-cooperative moving obstacles (and how to compensate if their trajectories are uncertain) is still a very interesting and open research question that we are currently working on. 

\bibliography{mybib}
\end{document}